\documentclass[10pt,twocolumn,letterpaper]{article}

\usepackage{cvpr}              
\usepackage{amsmath}
\usepackage{multirow}

\usepackage[dvipsnames]{xcolor}

\definecolor{cvprblue}{rgb}{0.21,0.49,0.74}
\usepackage[pagebackref,breaklinks,colorlinks,citecolor=cvprblue]{hyperref}

\def\paperID{8 } 
\def\confName{CVPR}
\def\confYear{2024}

\title{DemosaicFormer: Coarse-to-Fine Demosaicing Network for HybridEVS Camera}

\author{Senyan Xu$^\ast$, Zhijing Sun\thanks{ Co-first authors, $^\dagger$corresponding author.}, Jiaying Zhu, Yurui Zhu, Xueyang Fu, Zheng-Jun Zha$^\dagger$\\
University of Science and Technology of China\\
{\tt\small \{syxu, sunzhijing, zhujy53, zyr\}@mail.ustc.edu.cn,  \{xyfu, zhazj\}@ustc.edu.cn}
}

\begin{document}
\maketitle
\begin{abstract}
Hybrid Event-Based Vision Sensor (HybridEVS) is a novel sensor integrating traditional frame-based and event-based sensors, offering substantial benefits for applications requiring low-light, high dynamic range, and low-latency environments,  such as smartphones and wearable devices.  Despite its potential, the lack of Image signal processing (ISP) pipeline specifically designed for HybridEVS poses a significant challenge. To address this challenge, in this study, we propose a coarse-to-fine framework named DemosaicFormer which comprises coarse demosaicing and pixel correction. Coarse demosaicing network is designed to produce a preliminary high-quality estimate of the RGB image from the HybridEVS raw data while the pixel correction network enhances the performance of image restoration and mitigates the impact of defective pixels. Our key innovation is the design of a Multi-Scale Gating Module (MSGM)  applying the integration of cross-scale features, which allows feature information to flow between different scales. Additionally, the adoption of progressive training and data augmentation strategies further improves model's robustness and effectiveness. Experimental results show superior performance against the existing methods both qualitatively and visually, and our DemosaicFormer achieves the best performance in terms of all the evaluation metrics in the MIPI 2024 challenge on Demosaic for Hybridevs Camera. The code
is available at \href{https://github.com/QUEAHREN/DemosaicFormer}{ this repository}.

\end{abstract}    
\section{Introduction}
\label{sec:intro}
\begin{figure}
    \centering
    \includegraphics[width=0.88\linewidth]{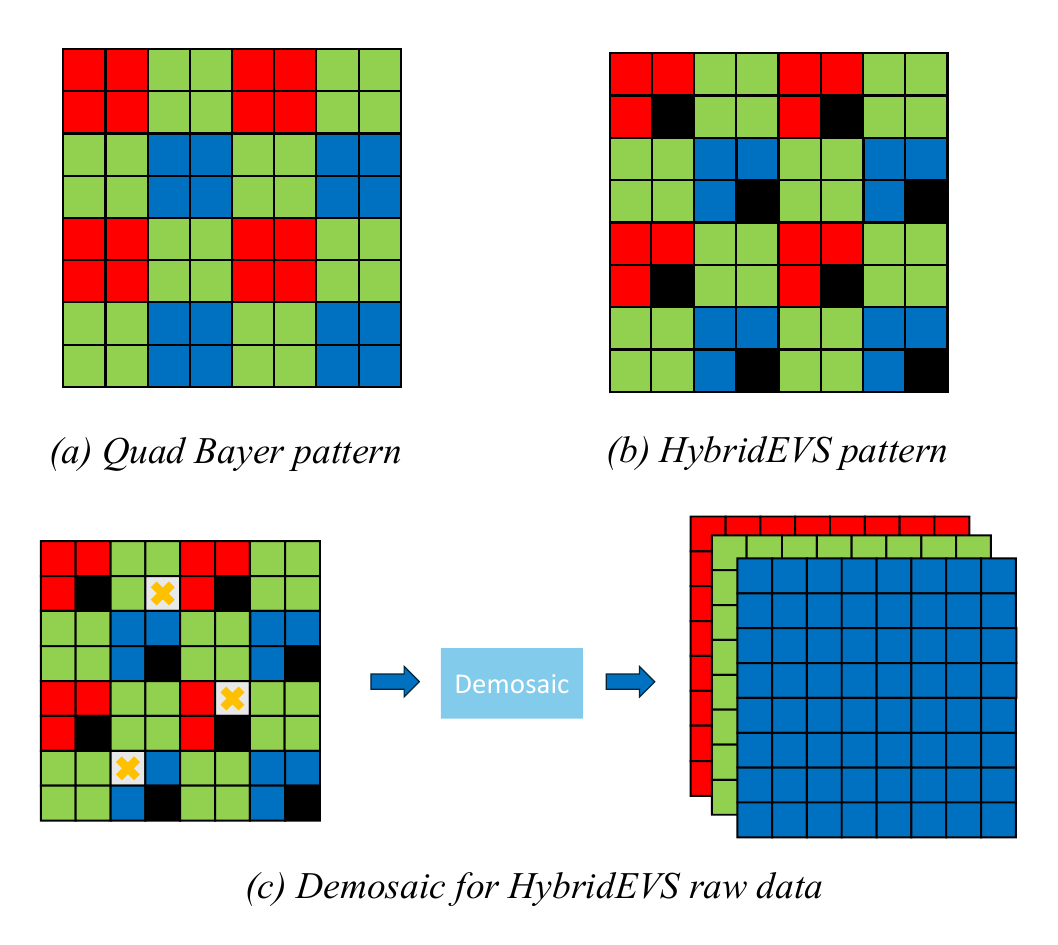}
    \caption{Illustration of two different patterns and demosacing task. (a) Quad Bayer pattern. (b) HybridEVS pattern.  (c) Demosaic for HybridEVS Camera task refers to the conversion of HybridEVS pattern raw data into RGB images.}
    \label{fig:pattern}
\end{figure}

Event-Based Vision Sensor (EVS) detects luminance changes asynchronously and will output event data immediately, which has the advantages of low power consumption and high sensitivity, and is suitable for capturing high dynamic range visual information without blurring. However, the inability to capture color information greatly limits the application scope of event cameras. Hybrid Event-based Vision Sensor (HybridEVS) \cite{hybridevs2023sony} is a novel hybrid sensor formed by combining traditional frame-based sensor and event-based sensor. It combines the advantages of these sensors, offering high temporal resolution, low latency, and exceptional dynamic range while still capturing color information with higher Signal-to-Noise Ratio (SNR). Compared to traditional sensors, HybridEVS can perform better in a greater range of applications because of its hybrid design.  
Quad Bayer pattern, as shown in Fig.~\ref{fig:pattern}(a) is a common type of pattern widely employed in smartphone cameras due to its ability to obtain high-quality images under low light secnary by averaging four pixels within a $2\times2$ neighborhood. While signal-to-noise ratio (SNR) is improved in the binning mode, the spatial resolution is reduced as a tradeoff. Defect pixels are flaws caused by the sensor's manufacturing process, where certain pixel values are inaccurate during the photoelectric conversion process.

\begin{figure*}[ht]
    \centering
    \includegraphics[width=0.88\linewidth]{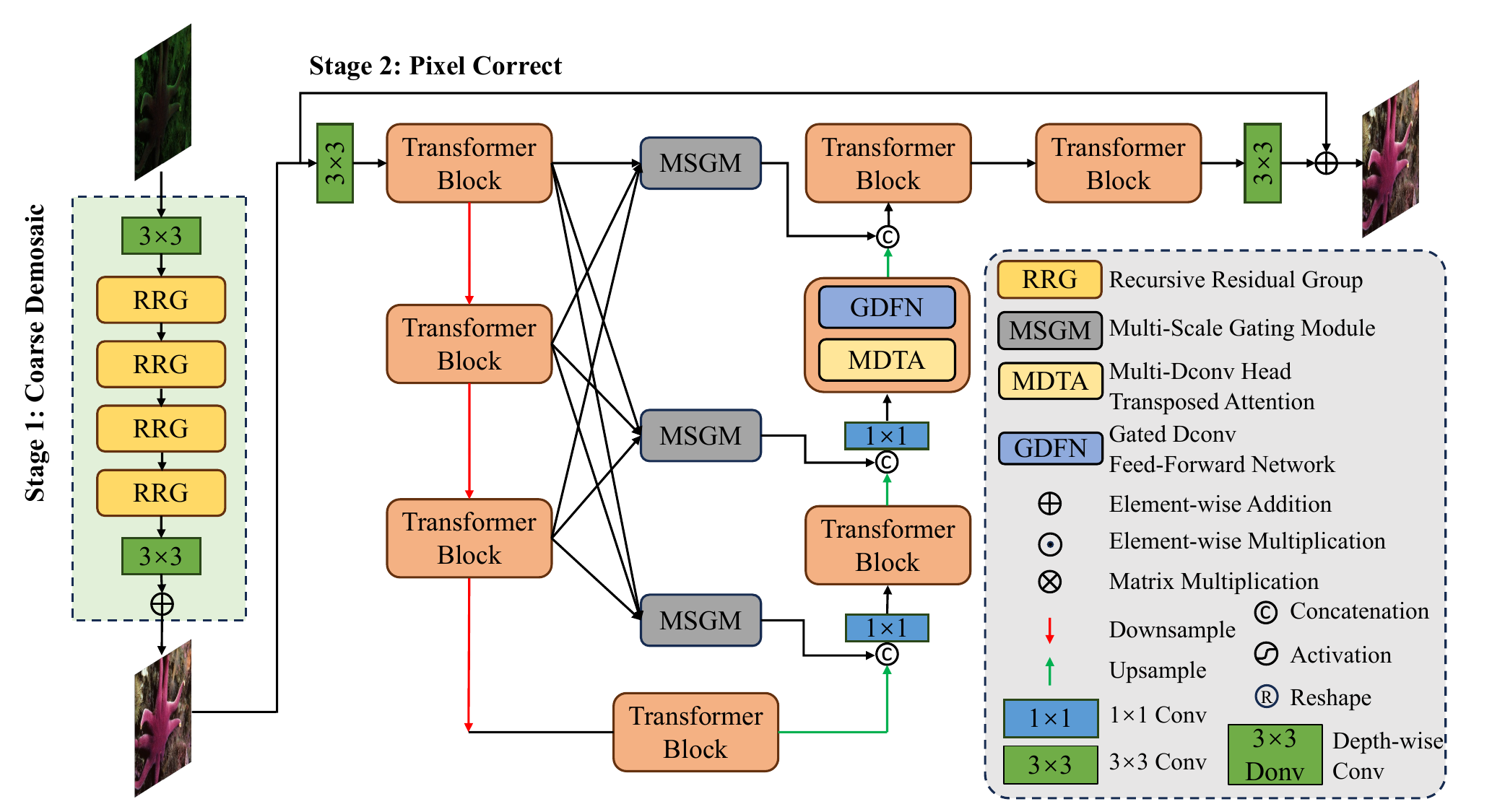}
    \caption{The architecture of our proposed DemosaicFormer to demosaic the raw data captured by HybridEVS cameras.}
    \label{fig:arch_1}
\end{figure*}

HybridEVS pattern, as shown in Fig.~\ref{fig:pattern}(b), is based on Quad Bayer pattern which replaces two normal pixels in the $4\times4$ pattern by event pixels (represented by black pixels). However, conventional general-purpose methods face challenges when demosacing for HybridEVS raw data. Since  Quad Bayer pattern sacrifices spatial resolution and event pixels can not record color information, demosaicing for HybridEVS raw data has less spatial and color information than demosaicing for regular raw data. On the other hand, as with any sensor, defect pixels can occur. Therefore, with the HybridEVS pattern, identifying and correcting these pixels is more challenging.

To address this challenge, we propose a coarse-to-fine framework named DemosaicFormer which comprises a coarse demosaicing network and a pixel correction network. For the coarse demosaicing stage, in order to produce a preliminary high-quality estimate of the RGB image from the HybridEVS raw data, we introduce Recursive Residual Group (RRG) \cite{zamir2020cycleisp} which employs multiple Dual Attention Blocks (DABs) to refine the feature representation progressively.  For pixel correction stage, aiming to enhance the performance of image restoration and mitigate the impact of defective pixels, we introduce the Transformer Block which consists of Multi-Dconv Head Transposed Attention (MDTA) and Gated-Dconv Feed-Forward Network (GDFN). Our key innovation is the design of a novel Multi-Scale Gating Module (MSGM) applying the integration of cross-scale features, which allows feature information to flow between different scales.The main contributions of our paper are summarized as follows:

\begin{enumerate}
    \item[-] We present a novel coarse-to-fine framework (called DemosaicFormer) to demosaic for HybridEVS raw images with defect pixels which decomposes the task into two sub-tasks: coarse demosaicing and pixel correction. 
    \item[-] We devise the Multi-Scale Gating Module (MSGM) to enhance the network by improving the interaction of feature information flow among different scales.
    \item[-] Experimental results show that the proposed method significantly outperforms other exited solutions. In the MIPI-challenge 2024 Demosaic for HybridEVS Camera track, our DemosaicFormer achieves first place in terms of all the evaluation scores (PSNR, SSIM) and outperforms the others by a large margin.
\end{enumerate}

\section{Related Work}
\begin{figure*}
    \centering
    \includegraphics[width=0.88\linewidth]{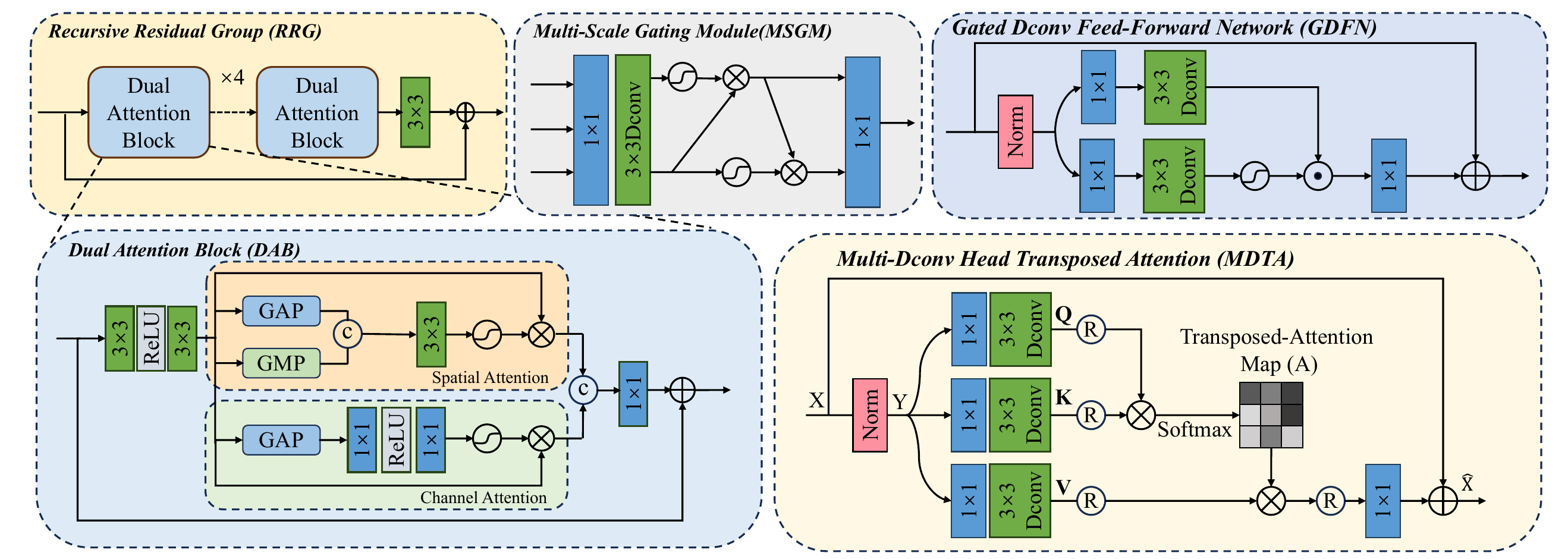}
    \caption{The structures of sub-modules in the main architecture.}
    \label{fig:arch_2}
\end{figure*}

\subsection{Image Signal Processing Pipeline}
Image signal processing (ISP) pipeline is a series of processing steps in digital image processing that are used to convert raw images obtained from cameras or other image acquisition devices into final usable images. This pipeline typically consists of multiple stages, each performing specific image processing tasks to improve image quality, enhance specific image features, or prepare the image for subsequent processing or display. ISP includes a series of processing algorithms that process raw images to obtain RGB images, such as demosaic, denoising, gamma correction, etc. With the development of deep neural networks (DNN), many studies \cite{deepisp2018,liang2021cameranet,ignatov2020replacing} use DNN to directly replace the main processing flow of ISP and convert raw images into RGB images end-to-end. CycleISP \cite{zamir2020cycleisp} uses a cyclic approach to construct a noise data set of real scenes, modeling the camera imaging pipeline in both forward (RGB2RAW) and reverse (RAW2RGB) directions. 

\subsection{Deep Learning for Image Restoration}

Image restoration aims to recover its clean counterparts from a degraded image. A popular scheme is to use CNN structures to learn efficient models to capture local features of images and learn generalizable image priors. CNNs have been widely used in various image restoration tasks, including image denoising \cite{cheng2021nbnet,yue2020dual}, demosaicing \cite{demosacicnet2022zhang,liu2018demoir}, and super-resolution \cite{dasr2021wang,zhang2018imagesr}. 
Chen et al. \cite{nafnet} used multiplication to replace or delete unnecessary activation functions such as Sigmoid, ReLU, GELU, and Softmax, and derived a nonlinear activation free network called NAFNet. Zhu et al. \cite{zhu2023enhanced} proposed ECFNet to effectively restore UDC images which takes multi-scale images as input. MIRNet \cite{mirnet} is a novel architecture that learns a rich set of features incorporating contextual information from multiple scales while maintaining high resolution.

After the Transformer model shined in the field of natural language processing, Vision Transformer (ViT) \cite{dosovitskiy2020imagevit} has also been extensively explored in high-level visual tasks, such as object detection \cite{detr,zhu2020deformable}, image segmentation \cite{xie2021segformer,zheng2021setr}, etc. Transformer has the ability to capture long-range dependencies between image patches and adapt to given input content. Due to these characteristics, Transformer is also used in the field of image restoration \cite{zhou2023srformer,liang2021swinir,wang2022uformer}. ShuffleFormer \cite{shuffleformer} proposes a local window Transformer based on a random shuffling strategy to model non-local interactions with linear complexity. Restormer \cite{zamir2022restormer} proposes an efficient Transformer-based model.

\subsection{HybridEVS Visions}

Event-Based Vision Sensor Camera has the advantages of low power consumption and high sensitivity, and is suitable for capturing high dynamic range visual information without blurring. There have been related works using Deep Neural Network (DNN) with RGB and event data for effective image enhancement (such as deblurring and video frame interpolation) \cite{event2020deblur,event2022timelens}. But these image processing techniques require equivalent RGB characteristics to advanced mobile RGB sensors, as well as alignment of focus between RGB and event pixels on the sensor. Based on this, Kodama et al. \cite{hybridevs2023sony} proposed the Hybrid Event-Based Vision Sensor, which can achieve image enhancement of mixed data in mobile application processors. However, the manufacturing process of the sensor will cause defects, and there will also be some inaccurate pixel values during the photoelectric conversion process, resulting in the appearance of defective pixels. Currently, the reconstruction of HybridEVS raw data containing event pixels and defective pixels into RGB images is less explored.
\section{Method}

Our proposed DemosaicFormer is built with two-stage cascade framework, which gradually generates desired high-quality results for Hybridevs camera in a coarse-to-fine manner. As shown in Fig.~\ref{fig:arch_1}, the proposed framework consists of coarse demosaicing and pixel correction network, which is based on the CycleISP \cite{zamir2020cycleisp} and Restormer \cite{zamir2022restormer} respectively. Different from these approaches, our two-stage framework can decompose the complex task into individual sub-tasks which can increase each network’s learning ability and make the whole framework easier to converge. Furthermore, we devise the Multi-Scale Gating Module (MSGM) to transfer the feature information flow among the Transformer Blocks of cross scales. Following this, we present detailed explanations of our pipeline and the key components encompassed within proposed approach.
\begin{table}[]
\caption{Quantitative comparisons of methods on the official testing datasets of the MIPI-challenge 2024 Demosaic for Hybridevs Camera track. The best and
the second results are boldfaced and underlined, respectively. \label{challenge_res}}
\begin{tabular}{c|c|c c}
\toprule
\multirow{2}{*}{\textbf{Rank}}&\multirow{2}{*}{\textbf{Methods}} & \multicolumn{2}{c}{Metrics} \\ [0.5ex]
         ~ &~ & \textbf{PSNR}$\uparrow$ & \textbf{SSIM}$\uparrow$ \\
         
         \hline\hline
1    & DemosaicFormer(Ours)               & \textbf{44.8464} & \textbf{0.9854} \\
2    & 2nd                & \underline{44.6234} & \underline{0.9847} \\
3    & 3rd   & 44.4950  & 0.9845 \\
4    & 4th & 43.9564 & 0.9837 \\
5    & 5th                & 42.6508 & 0.9810 \\
6    & 6th            & 41.3279  & 0.9780 \\
7    & 7th             & 41.0737 & 0.9752 \\ 
\bottomrule
\end{tabular}
\end{table}

\subsection{Overall Pipeline}

In some learning ISP methods \cite{liang2021cameranet, ignatov2020replacing, deepisp2018}, defect pixel removal and demosaicing are often implemented in one stage due to the relatives between them. So we first feed the original raw image into the coarse demosaicing network to get an imperfect image in RGB space.  Then, the RGB image will go through the pixel correction network which gradually restores the corrupted RGB image in a coarse to fine manner. The second stage finally outputs a desired high-quality RGB image.

In detail, for coarse demosaicing stage, given a HybridEVS raw image of  $\mathbf{I}_{raw} \in \mathbb{R}^{H \times W \times 1}$, we extend it to RGB space $\mathbf{I}_{raw}^{RGB} \in \mathbb{R}^{H \times W \times 3}$, a coarse demosaicing network noted as $\mathbf{F}_{cd}$ is employed to simply eliminate the defect pixels and restore the raw image to RGB space $\mathbf{I}_{rest}^{RGB}$. 
\begin{equation}
    \mathbf{I}_{rest}^{RGB} = {F}_{cd}(Extend(\mathbf{I}_{raw}))    
\label{311}
\end{equation}
After that, $\mathbf{I}_{rest}^{RGB}$ is taken as the input of pixel correction stage and a pixel correction network noted as $\mathbf{F}_{pc}$  is adopted to correct pixel and refine the imperfect image.
\begin{equation}
    \mathbf{I}_{output}^{RGB} = {F}_{pc}(\mathbf{I}_{rest}^{RGB})   
\label{312}
\end{equation}
Finally, we get the desired images $\mathbf{I}_{output}^{RGB} \in \mathbb{R}^{H \times W \times 3}$ . The whole two-stage framework can be formulated as:
\begin{equation}
    \mathbf{I}_{output}^{RGB} = {F}_{pc}({F}_{cd}(Extend(\mathbf{I}_{raw}), \theta_{cd}),\theta_{pc})       
\label{313}
\end{equation}
Here $\theta_{cd}$, $\theta_{pc}$ denote the learnable parameters in $\mathbf{F}_{cd}$ and $\mathbf{F}_{pc}$. 
By decomposing complex demosaic tasks, our DomisaicFormer achieves outstanding results.

\subsection{Coarse Demosaicing Network}
The Coarse Demosaicing Network aims to produce a preliminary high-quality estimate of the RGB image from the raw data. Inspired by  \cite{he2016deep,ren2019progressive,zhang2018multi}, we introduce Recursive residual group (RRG) \cite{zamir2020cycleisp} which employs multiple Dual Attention Blocks (DABs) to refine the feature representation progressively. As shown in Fig.~\ref{fig:arch_2},  the DAB is a comprehensive attention unit within the RRG that utilizes both spatial\cite{woo2018cbam} and channel\cite{hu2018squeeze} attention mechanisms. The overall process of DAB is:
\begin{equation}
    {T}_{DAB} = {T}_{in} + Conv(Concat([CA(U),SA(U)]))     
\label{321}
\end{equation}
where $U\in\mathbb{R}^{H \times W \times C}$ denotes tensors of features maps obtained by applying two  $3\times3$ conv layers on input tensor $T_{in} \in \mathbb{R}^{H \times W \times C}$, $Conv(\cdot)$ is the last $1\times1$ conv layer.
\begin{figure}
    \centering
    \includegraphics[width=\linewidth]{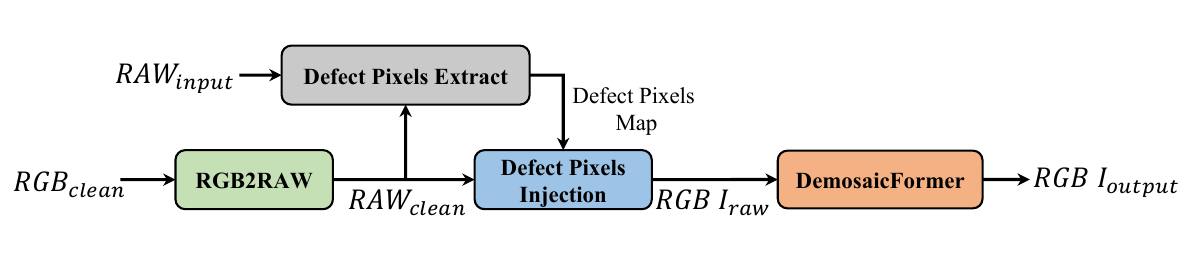}
    \caption{Train model using synthesized augmented data.}
    \label{fig:initail}
\end{figure}

\begin{table*}[]
    \centering
    \caption{Quantitative comparisons of methods on the official validation datasets of the MIPI-challenge 2024 Demosaic for Hybridevs Camera track. The MACs and FLOPs is computed using a $128\times128$ image as input by calflops tool. The best and
the second results are boldfaced and underlined, respectively.  }
    \begin{tabular}{l|c|c|c|c c}
    \toprule
         \multirow{2}{*}{\textbf{Methods}}  & \multirow{2}{*}{\textbf{\#Params (M)}}  &\multirow{2}{*}{\textbf{MACs (G)}}  &\multirow{2}{*}{\textbf{FLOPs (G)}}  &\multicolumn{2}{c}{Metrics} \\ [0.5ex]
          ~ & ~ &~ &~&\textbf{PSNR}$\uparrow$ & \textbf{SSIM}$\uparrow$ \\
         \hline\hline
         CycleISP\cite{zamir2020cycleisp}  & 2.8 &46.0 &93.4 &41.32 & 0.98\\
         MIMO-UNet\cite{cho2021mimo} & 8.9 & 21.1 &41.2&40.75 & 0.98\\
         MIMO-UNet${}^*$\cite{cho2021mimo} &8.9  &21.7  & 41.7 &41.27 & 0.98\\
         ECFNet\cite{zhu2023enhanced} &9.1 &21.7 & 42.5 &41.45 & 0.98\\
         NAFNet\cite{nafnet} & 67.8 &15.8 & 31.6&41.19  &0.98\\
         MIRNet\cite{mirnet}& 5.9 &34.9 & 70.0 &40.92& 0.98\\
         Restormer\cite{zamir2022restormer}& 26.1 &35.2&70.6&\underline{41.73} & \underline{0.98}\\
         ShuffleFormer\cite{shuffleformer} & 50.6 &20.7 &41.6 &41.70 & 0.98\\
         DemosaicFormer(Ours)  & 30.3  &85.1 &171.5  &\textbf{42.01} & \textbf{0.98}\\
    \bottomrule
    \end{tabular}
    \label{challenge_valres}
\end{table*}

\subsection{Pixel Correction Network}
The output of coarse demosacing stage still suffers from the impact of defect pixels because the first stage can't perfectly tackle joint demosaic and defect pixels removal tasks. Pixel Correction Network is aimed to enhance the performance of image restoration and mitigate the impact of defective pixels.  Existing CNN-based image restoration methods have achieved impressive results \cite{cho2021mimo,mirnet,nafnet,zhu2023enhanced}. However, these approaches exhibit shortcomings in capturing long-range dependencies and non-local similarities. In contrast, Transformer methods have shown exceptional ability over the past few years with great performance. However, directly applying a conventional Transformer has more computational overhead which comes from the self-attention layer. Moreover, regular Transformer architectures always overlook the integration of cross-scale features, which is crucial for effective image restoration. To address this problem, inspired by \cite{zamir2022restormer}, we introduce the Transformer Block which consists of Multi-Dconv Head Transposed Attention (MDTA), Gated-Dconv Feed-Forward Network (GDFN) and Multi-Scale Gating Module (MSGM). 

\textbf{Multi-Dconv Head Transposed Attention(MDTA)}, shown in Fig.~\ref{fig:arch_2}  has linear complexity implemented by applying conventional SA \cite{vaswani2017attention} across channels dimension which is the key design of MDTA. As another important component of MDTA,  depth-wise convolutions generate the global attention map emphasizing on the local context before computing attention.

\textbf{Gated-Dconv Feed-Forward Network(GDFN)}, shown in Fig.~\ref{fig:arch_2}, is utilised to transform features after MDTA, which is different from the regular feed-forward network(FN)\cite{dosovitskiy2020imagevit}. To improve representation learning, gating mechanism and depthwise convolutions are applied in GDFN.The gating mechanism is structured as the Hadamard product (element-wise multiplication) of two parallel pathways consisting of linear transformation layers. Similar to MDTA, all pathways include $3\times3$ depth-wise convolutions to encode information from spatially neighboring pixel positions, useful for learning local image structure for effective restoration.   One of these pathways is activated with the Gaussian Error Linear Unit (GELU)\cite{hendrycks2016gaussian} .

\textbf{Multi-Scale Gating Module (MSGM)} Inspired by ResNet, some methods supplement the original features in the encoder to the decoder through skip connection. This can reduce the difficulty of network optimization and improve network performance. In some cases, features are even transferred across scales, feeding features from the encoder into different scales of the decoder.
In this paper, inspired by NAFNet \cite{nafnet}, we furthermore introduce a simple gating mechanism into cross-scale feature fusion increasing the nonlinearity of fusion. Based on the gating mechanism, we can extract the features needed by different scale decoders which improves the correction effect of the network. Specifically, as shown in Fig.~\ref{fig:arch_2}, our Multi-Scale Gating Module (MCGM) up-samples or down-samples the features at different scales according to the required shape of the module output, then concatenates them at the channel dimension and adjusts the number of channels using 1$\times$1 convolution. Inspired by the simple gate in NAFNet, we divide them into two equal parts for 3$\times$3 depth-wise convolution. Each feature is multiplied by the sigmoid change of the other feature, and finally the two parts of the features are transformed into the required enhancement features using 1$\times$1 convolution. Formally, the MCGM at the shallowest scale can be presented as
\begin{equation}
    \begin{split}
        F &= \mathit {Conv}(\mathit {Concat}([(TB_{1}^{out}), (TB_{2}^{out})^{\uparrow} , (TB_{3}^{out})^{\uparrow}])), \\
    F_{1}, F_{2} &= \mathit {Split}(DConv(F)), \\
    F_{1} &= Sigmoid(F_{1})\times F_{2}, \\
    F_{2} &= Sigmoid(F_{2})\times F_{1}, \\
    F_{e} &= Conv(Concat([F_{1},F_{2}])),
    \end{split}
\label{equ5}
\end{equation}
where $TB_{i}^{out}, i=1,2,3$ denotes the output of the $n^{th}$ scale transformer block, $\uparrow$ denotes the up-sampling operation, $Concat(\cdot)$ denotes the concatenation operation along the channel dimension, $Dconv(\cdot)$ denotes the depth-wise convolution, $Split(\cdot)$ denotes the chunk operation.

\begin{table}[]
    \centering
    \caption{ Quantitative comparisons of different training objects. The best result is boldfaced.  }
    \renewcommand{\arraystretch}{1.1}
    \begin{tabular}{c|l|c}
    \toprule
         \textbf{Model}& \textbf{Training Description}  & \textbf{PSNR}$\uparrow$  \\
         
         \hline\hline
         A& \space\space Indiv. Train.  \& Joint FT \space \space\space\space\space\space\space\space\space\space\space\space\space\space\space\space\space\space& 41.99  \\

        B&\space\space Joint Train. w. Ext. Sup.  & 40.76  \\
        C&\space\space Joint Train. (default)   & \textbf{42.01 } \\

    \bottomrule
    \end{tabular}
    \label{exp_object}
\end{table}
\begin{table*}[]
    \centering
    \caption{ Ablation study of the  Training Strategies. The best and
the second results are boldfaced and underlined, respectively.  }
    \begin{tabular}{l|c|c|c|c|c}
    \toprule
         \multirow{2}{*}{\textbf{ }}  & \multirow{2}{*}{\textbf{Progressive Training}} & \multirow{2}{*}{\textbf{Data Augmentation}} & \multirow{2}{*}{\textbf{Finetune Stage}} & \multicolumn{2}{c}{\textbf{PSNR}$\uparrow$ } \\
         ~ & ~ & ~ & ~ &\textbf{VAL SET}&\textbf{TEST SET} \\
         \hline\hline
         DemosaicFormer  & ~ & ~ & ~ & 42.01 & - \\
         DemosaicFormer & ~ & 50\% Prob. & ~ & 42.39  & 42.61\\
         DemosaicFormer  & ~ & \checkmark & ~ & 43.10  & 42.54 \\
         DemosaicFormer-s1 & \checkmark & \checkmark & ~ & \underline{43.17}  & \underline{42.63} \\
         DemosaicFormer-s2  & \checkmark & \checkmark & \checkmark &\textbf{43.26 }   & \textbf{42.98 } \\
    \bottomrule
    \end{tabular}
    \label{abl_training}
\end{table*}
\subsection{Joint Training of DemosaicFormer}

Given the intrinsic interdependence of coarse demosaicing and pixel correction, it is impractical to disentangle them completely into separate subtasks. Hence, in our DemosaicFormer, we employ a joint training approach which will be discussed in Section \ref{training}. We utilize $\ell_1$ loss, which is widely used in many image restoration and enhancement tasks\cite{zhu2023enhanced,zamir2022restormer,nafnet,mirnet,zamir2020cycleisp,liang2021cameranet}. The loss function for the joint optimization is:
\begin{equation}
\mathcal{L}_{1}(I,\hat{I})=\frac{1}{N} \sum_{p \in P}|I_(p) - \hat{I}_(p)|,
\end{equation}
where $p$ is the index of the pixel and $P$ is the patch; $I$ and $\hat{I}$ represent the ground-truth and restored result by our DemosaicFormer with $N$ pixels, respectively.

\begin{table}[]
    \centering
    \caption{ Ablation study of the connection manner in the different level. The best and
the second results are boldfaced and underlined, respectively.  }
    \begin{tabular}{c|l|c}
    \toprule
         \textbf{ Level}&\textbf{ Connection Manner}  & \textbf{PSNR}$\uparrow$  \\
         \hline\hline
         \multirow{3}{*}{Arch}&Pixel Correct First  & \underline{42.92}  \\
         ~ &Coarse Demosaic First(default)  &  \textbf{43.10} \\
         ~ & Parallel Connection  & 42.85  \\
         \hline 
         \multirow{3}{*}{Block} &Simple Concatenation  & 42.93  \\
         ~ &Single Gating Fusion  & \underline{42.99}   \\
         ~ &Multi-Scale Gating Module(default) & \textbf{43.10} \\
    \bottomrule
    \end{tabular}
    \label{abl_connection}
\end{table}

\section{Experiments}

\subsection{Dataset}
We conduct the experiments strictly following the setting of the MIPI-challenge 2024 Demosaic for Hybridevs Camera track\cite{hybridevs2024mipi3}. The training data consists of 800 pairs of Hybridevs's input data and label result with a resolution of 2K. Both the input and label have the same spatial resolution. The input is of 10bits in the “.bin” format and ranges from [0, 1023], and the corresponding ground truth is of 8bits in the “.png” format. The validation and testing sets consist of 50 images each, and each set contains images of varying resolutions. In the testing set, the resolution of images is not fixed, ranging from $1280\times720$ to $5760\times5760$. Note that the ground truth data corresponding to the validation and testing dataset is not publicly available. 

\textbf{Data augmentation}. Due to our inability to accurately model defective pixels, inspired by \cite{zamir2020cycleisp}, we extract the defect pixels map from the training data of the challenge to generate more diverse and realistic inputs. As shown in Fig.~\ref{fig:initail}, at the training phase, we randomly rotate and flip ground-truth images($RGB_{clean}$) of training split, then sample them according to HybridEVS pattern, and randomly cover the sampled images with defect pixels map. The augmentation technology is applied at the initial training of our proposed approach for improving the model's generalization and robustness. Note that the models trained with different data augmentation strategies are different, as seeing in the section \ref{Implementation}.


\begin{figure*}[ht]
    \centering
    \includegraphics[width=0.8\linewidth]{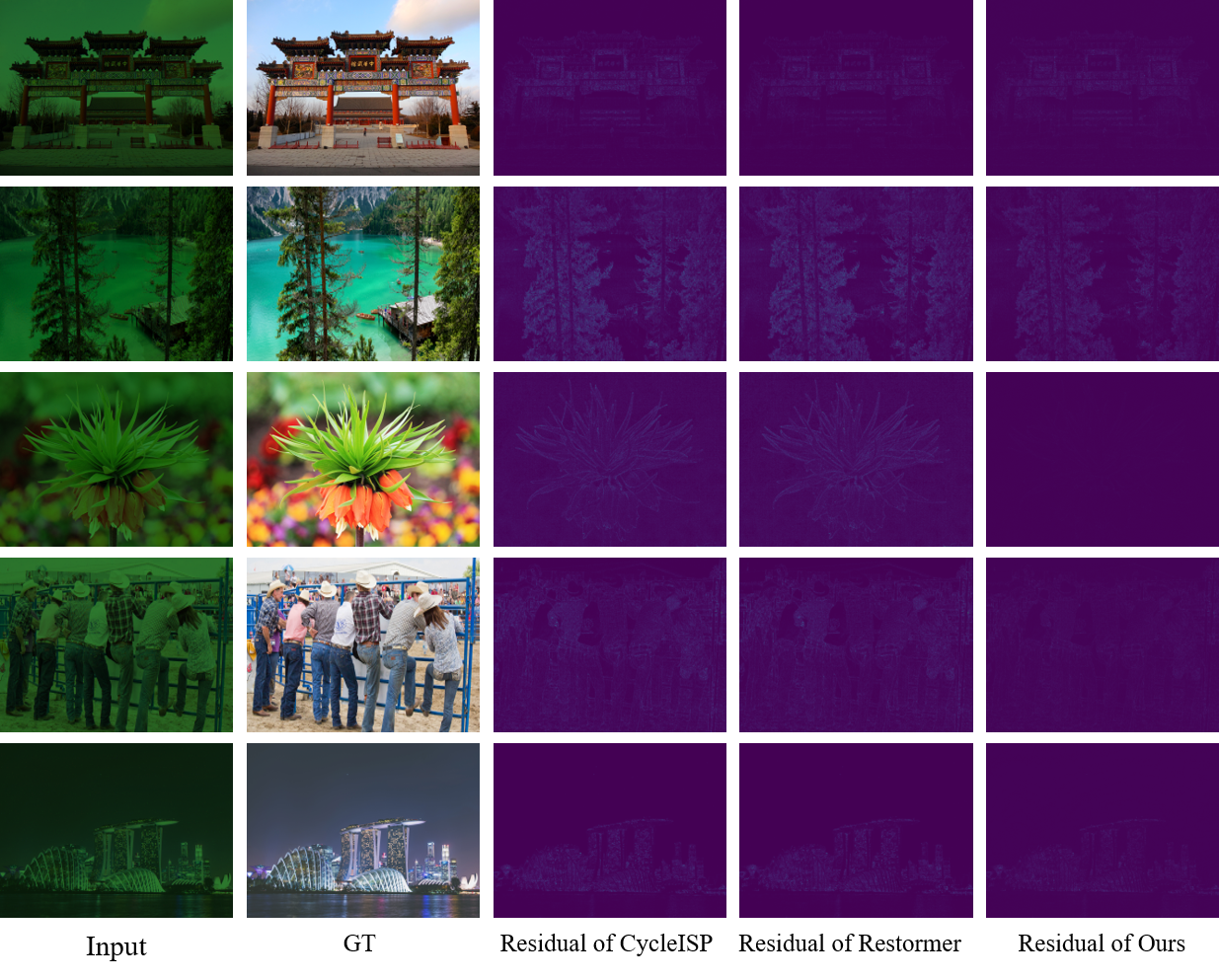}
    \caption{Visual comparison results of Demosaic for Hybridevs Camera on the evaluation dataset of MIPI-challenge 2024 track. Note that brighter means bigger error.}
    \label{fig:residual_map}
\end{figure*}

\begin{figure*}[ht]
    \centering
    \includegraphics[width=0.8\linewidth]{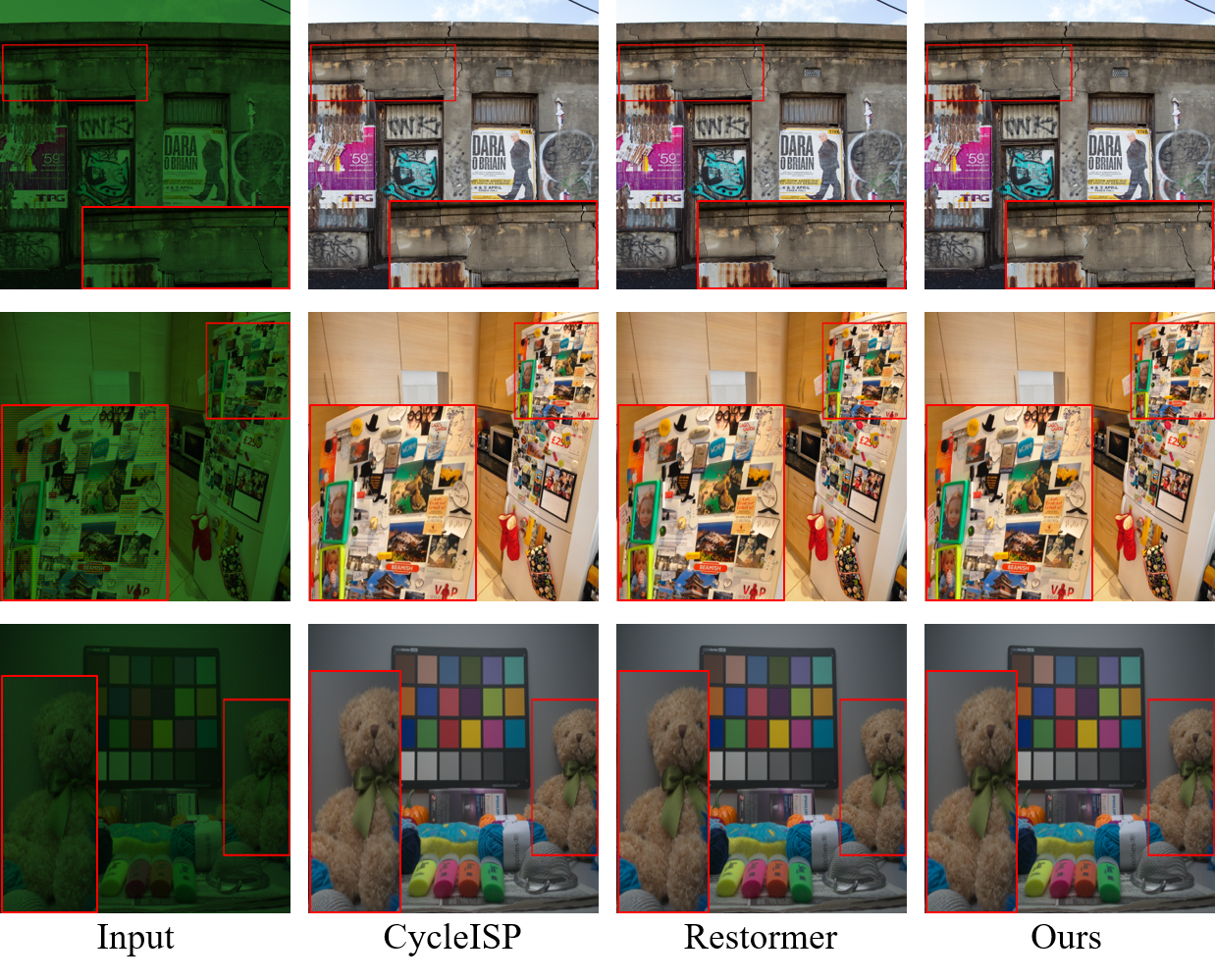}
    \caption{Visual comparison results of Demosaic for Hybridevs Camera on the testing dataset of MIPI-challenge 2024 track.}
    \label{fig:detail}
\end{figure*}

\subsection{Implementation Details}
\label{Implementation}
We implement our proposed network via the PyTorch 1.8 platform. Adam optimizer with parameters $\beta_1 = 0.9$ and $\beta_2 = 0.99$ is adopted to optimize our network. Additionally, motivated by \cite{zamir2022restormer}, we introduce the progressive training strategy. The training phase of our network could be divided into two stages:

(1) \textbf{Initial training of DemosaicFormer}. We use a progressive training strategy at first. We start training with patch size $80\times80$ and batch size 84 for 58K iterations.  The patch size and batch size pairs are updated to $[(128^2,30), (160^2,18), (192^2,12)]$ at iterations [ 36K, 24K, 24K]. The initial learning rate is $5 \times 10^{-4}$ and remains unchanged when patch size is 80. Later the learning rate changes with Cosine Annealing scheme to $1 \times 10^{-7}$. For data augmentation, we use our data augmentation mentioned above. The first stage is performed on the NVIDIA 4090 device. We obtain the best model at this stage as the initialization of the second stage.

(2) \textbf{Fine-tuning DemosaicFormer}.We start training with patch size  $192 \times 192$ and batch size 12. The initial learning rate is $1 \times 10^{-4}$ and changes with Cosine Annealing scheme to $1 \times 10^{-7}$, including 20K iterations in total. We use the entire training data from the challenge without any data augmentation technologies. Exponential Moving Average (EMA) is applied for the dynamic adjustment of model parameters. The second stage is performed on the NVIDIA 4090 device. 

To better distinguish between the model results, we label the two stages as DemosaicFormer-s1 and DemosaicFormer-s2, respectively.

\subsection{Evaluation Metrics}
We employ two reference-based metrics which are widely applied in similar tasks\cite{liang2021cameranet,zamir2022restormer,nafnet,mirnet,cho2021mimo,liang2021swinir}, to assess the efficacy of our method: Peak Signal-to-Noise Ratio (PSNR), the structural similarity (SSIM) \cite{ssim}.  Higher values of PSNR and SSIM indicate better performance in image restoration tasks. Note that due to the evaluation settings of the challenge, we are unable to obtain the exact SSIM value, but it does not affect the ordering of SSIM.

\subsection{Comparations}
\label{training}
Table \ref{challenge_res} presents a comprehensive comparison of various solutions on the MIPI-challenge 2024 Demosaic for Hybridevs Camera track. Evidently, our approach outperforms all others across all evaluation metrics on the official testing datasets, showcasing superior performance. Specifically, our method achieves a remarkable improvement, surpassing the second-place method by 0.2230 dB in PSNR. 

Besides, in Table \ref{challenge_valres}, we demonstrate comparable performance methods on the official validation datasets when compared to some ISP methods and general image restoration methods. For a fair comparison, note that all methods utilize HybridEVS's raw data expanded into RGB space as input without any data augmentation techniques.  Our method consistently demonstrates outstanding performance.  Compared to the method Restormer and ShuffleFormer, we obtain 0.28dB and 0.31dB gain in PSNR.
Furthermore, in Fig. \ref{fig:residual_map},\ref{fig:detail}, to more intuitively show our excellent performance, we generate the residual map representing the disparity between the predicted output and the ground truth.    The comparison clearly demonstrates that our technology produces superior visual results and outperforms others in terms of visual quality. Especially, our method reconstructs finer details more effectively and shows less departure from the ground truth,  demonstrating its efficiency in image restoration.

For training objects, Table \ref{exp_object} presents the results of employing various training objects for DemosaicFormer. Model A is two-phase training procedure where Coarse Demosaic Network is initially trained to convert raw data into RGB images, followed by joint finetuning, which extends training duration. Model B denotes joint training with extra constraint loss at Coarse Demosaic stage. Model C, in contrast, represents joint training  devoid of any extra constraints. The comparison clearly demonstrates that directly joint training can produce better results with less time durtation. Specifically, the model is encouraged to jointly optimize both the demosaicing task and any auxiliary tasks, thereby leveraging the interconnectedness inherent between two stages.

\subsection{Ablation Study}
We conduct plenty of ablation experiments to verify the effect of each component of our method. Note that in the ablation study with the absence of other annotations, we train the model with our data augmentation technology and without progressive training manner for convenience.

\textbf{Effects of the Connection Manner.} As shown in Table \ref{abl_connection}, we verify the validity of the DemosaicFormer connection manner at different levels, including the sequential choice of the two-stage network(arch-level) and the effectiveness of the MSGM module(block-level).  In connection manner of arch-level, we compare the performance of the model using different two-stage connection approaches which include exchanging the order of coarse demosaicing and pixel correction and processing the two branches in parallel. It is evident that coarse demosaicing before the pixel correction results in significant performance gains. Because of the sparsity nature of defect pixels, the initial demosaicing process is not significantly affected, while also providing more detailed color information for the post-processing.     Parallel processing causes degraded performance by disrupting the progressive processing flow created by cascading. 

Furthermore, in connection manner of block-level, effectiveness of the MSGM module is verified by replacing it with Simple Concatentation and Single Gating Fusion. The MSGM module incorporates multi-scale feature information and adaptively selects features based on the hierarchy of the output, obtaining 0.17dB gain in PSNR.

\textbf{Effects of the Training Strategies.} Following \cite{zamir2022restormer,zamir2020cycleisp}, we additionally adopt the progressive training strategy, various data augmentation strategies  
 and fine-tune model to enhance the model performance. As shown in Table \ref{abl_training}, the models trained at different stages are marked as DemosaicFormer-s1 and  DemosaicFormer-s2. Experiments show that training with progressively larger patches often results in higher gains in generalization performance. Our data augmentation technology greatly improves model's performance by increasing generalization and robustness. After initial training with progressive learning and data augmentation, fine-tuning the model on the original training set facilitates better adaptation to the real data distribution, obtaining 0.09dB and 0.35dB gain in PSNR on challenge official val set and test set, respectively.

\section{Conclusion}
In this paper, we present DemosaicFormer, an effective coarse-to-fine network for demosaicing HybridEVS's raw data. Built with a two-stage cascade framework comprising coarse demosaicing and pixel correction networks, DemosaicFormer decomposes the complex task into sub-tasks, and formulates Multi-Scale Gating Module(MSGM). Besides, the adoption of progressive training and data augmentation strategies further improves the model's robustness and effectiveness. DemosaicFormer achieves the best performance in terms of all the evaluation metrics in the MIPI-challenge 2024 Demosaic for Hybridevs Camera track.
\clearpage
{
    \small
    \bibliographystyle{ieeenat_fullname}
    \bibliography{main}

\begin{thebibliography}{36}
\providecommand{\natexlab}[1]{#1}
\providecommand{\url}[1]{\texttt{#1}}
\expandafter\ifx\csname urlstyle\endcsname\relax
  \providecommand{\doi}[1]{doi: #1}\else
  \providecommand{\doi}{doi: \begingroup \urlstyle{rm}\Url}\fi

\bibitem[Carion et~al.(2020)Carion, Massa, Synnaeve, Usunier, Kirillov, and Zagoruyko]{detr}
Nicolas Carion, Francisco Massa, Gabriel Synnaeve, Nicolas Usunier, Alexander Kirillov, and Sergey Zagoruyko.
\newblock End-to-end object detection with transformers.
\newblock In \emph{European conference on computer vision}, pages 213--229. Springer, 2020.

\bibitem[Chen et~al.(2022)Chen, Chu, Zhang, and Sun]{nafnet}
Liangyu Chen, Xiaojie Chu, Xiangyu Zhang, and Jian Sun.
\newblock Simple baselines for image restoration.
\newblock In \emph{European conference on computer vision}, pages 17--33. Springer, 2022.

\bibitem[Cheng et~al.(2021)Cheng, Wang, Huang, Liu, Fan, and Liu]{cheng2021nbnet}
Shen Cheng, Yuzhi Wang, Haibin Huang, Donghao Liu, Haoqiang Fan, and Shuaicheng Liu.
\newblock Nbnet: Noise basis learning for image denoising with subspace projection.
\newblock In \emph{Proceedings of the IEEE/CVF conference on computer vision and pattern recognition}, pages 4896--4906, 2021.

\bibitem[Cho et~al.(2021)Cho, Ji, Hong, Jung, and Ko]{cho2021mimo}
Sung-Jin Cho, Seo-Won Ji, Jun-Pyo Hong, Seung-Won Jung, and Sung-Jea Ko.
\newblock Rethinking coarse-to-fine approach in single image deblurring.
\newblock In \emph{Proceedings of the IEEE/CVF international conference on computer vision}, pages 4641--4650, 2021.

\bibitem[Dosovitskiy et~al.(2020)Dosovitskiy, Beyer, Kolesnikov, Weissenborn, Zhai, Unterthiner, Dehghani, Minderer, Heigold, Gelly, et~al.]{dosovitskiy2020imagevit}
Alexey Dosovitskiy, Lucas Beyer, Alexander Kolesnikov, Dirk Weissenborn, Xiaohua Zhai, Thomas Unterthiner, Mostafa Dehghani, Matthias Minderer, Georg Heigold, Sylvain Gelly, et~al.
\newblock An image is worth 16x16 words: Transformers for image recognition at scale.
\newblock \emph{arXiv preprint arXiv:2010.11929}, 2020.

\bibitem[He et~al.(2016)He, Zhang, Ren, and Sun]{he2016deep}
Kaiming He, Xiangyu Zhang, Shaoqing Ren, and Jian Sun.
\newblock Deep residual learning for image recognition.
\newblock In \emph{Proceedings of the IEEE conference on computer vision and pattern recognition}, pages 770--778, 2016.

\bibitem[Hendrycks and Gimpel(2016)]{hendrycks2016gaussian}
Dan Hendrycks and Kevin Gimpel.
\newblock Gaussian error linear units (gelus).
\newblock \emph{arXiv preprint arXiv:1606.08415}, 2016.

\bibitem[Hu et~al.(2018)Hu, Shen, and Sun]{hu2018squeeze}
Jie Hu, Li Shen, and Gang Sun.
\newblock Squeeze-and-excitation networks.
\newblock In \emph{Proceedings of the IEEE conference on computer vision and pattern recognition}, pages 7132--7141, 2018.

\bibitem[Ignatov et~al.(2020)Ignatov, Van~Gool, and Timofte]{ignatov2020replacing}
Andrey Ignatov, Luc Van~Gool, and Radu Timofte.
\newblock Replacing mobile camera isp with a single deep learning model.
\newblock In \emph{Proceedings of the IEEE/CVF conference on computer vision and pattern recognition workshops}, pages 536--537, 2020.

\bibitem[Jiang et~al.(2020)Jiang, Zhang, Zou, Ren, Lv, and Liu]{event2020deblur}
Zhe Jiang, Yu Zhang, Dongqing Zou, Jimmy Ren, Jiancheng Lv, and Yebin Liu.
\newblock Learning event-based motion deblurring.
\newblock In \emph{Proceedings of the IEEE/CVF Conference on Computer Vision and Pattern Recognition}, pages 3320--3329, 2020.

\bibitem[Kodama et~al.(2023)Kodama, Sato, Yorikado, Berner, Mizoguchi, Miyazaki, Tsukamoto, Matoba, Shinozaki, Niwa, et~al.]{hybridevs2023sony}
Kazutoshi Kodama, Yusuke Sato, Yuhi Yorikado, Raphael Berner, Kyoji Mizoguchi, Takahiro Miyazaki, Masahiro Tsukamoto, Yoshihisa Matoba, Hirotaka Shinozaki, Atsumi Niwa, et~al.
\newblock 1.22 $\mu$m 35.6 mpixel rgb hybrid event-based vision sensor with 4.88 $\mu$m-pitch event pixels and up to 10k event frame rate by adaptive control on event sparsity.
\newblock In \emph{2023 IEEE International Solid-State Circuits Conference (ISSCC)}, pages 92--94. IEEE, 2023.

\bibitem[Liang et~al.(2021{\natexlab{a}})Liang, Cao, Sun, Zhang, Van~Gool, and Timofte]{liang2021swinir}
Jingyun Liang, Jiezhang Cao, Guolei Sun, Kai Zhang, Luc Van~Gool, and Radu Timofte.
\newblock Swinir: Image restoration using swin transformer.
\newblock In \emph{Proceedings of the IEEE/CVF international conference on computer vision}, pages 1833--1844, 2021{\natexlab{a}}.

\bibitem[Liang et~al.(2021{\natexlab{b}})Liang, Cai, Cao, and Zhang]{liang2021cameranet}
Zhetong Liang, Jianrui Cai, Zisheng Cao, and Lei Zhang.
\newblock Cameranet: A two-stage framework for effective camera isp learning.
\newblock \emph{IEEE Transactions on Image Processing}, 30:\penalty0 2248--2262, 2021{\natexlab{b}}.

\bibitem[Liu et~al.(2018)Liu, Shu, and Wu]{liu2018demoir}
Bolin Liu, Xiao Shu, and Xiaolin Wu.
\newblock Demoir$\backslash$'eing of camera-captured screen images using deep convolutional neural network.
\newblock \emph{arXiv preprint arXiv:1804.03809}, 2018.

\bibitem[Ren et~al.(2019)Ren, Zuo, Hu, Zhu, and Meng]{ren2019progressive}
Dongwei Ren, Wangmeng Zuo, Qinghua Hu, Pengfei Zhu, and Deyu Meng.
\newblock Progressive image deraining networks: A better and simpler baseline.
\newblock In \emph{Proceedings of the IEEE/CVF conference on computer vision and pattern recognition}, pages 3937--3946, 2019.

\bibitem[Schwartz et~al.(2018)Schwartz, Giryes, and Bronstein]{deepisp2018}
Eli Schwartz, Raja Giryes, and Alex~M Bronstein.
\newblock Deepisp: Toward learning an end-to-end image processing pipeline.
\newblock \emph{IEEE Transactions on Image Processing}, 28\penalty0 (2):\penalty0 912--923, 2018.

\bibitem[Tulyakov et~al.(2022)Tulyakov, Bochicchio, Gehrig, Georgoulis, Li, and Scaramuzza]{event2022timelens}
Stepan Tulyakov, Alfredo Bochicchio, Daniel Gehrig, Stamatios Georgoulis, Yuanyou Li, and Davide Scaramuzza.
\newblock Time lens++: Event-based frame interpolation with parametric non-linear flow and multi-scale fusion.
\newblock In \emph{Proceedings of the IEEE/CVF Conference on Computer Vision and Pattern Recognition}, pages 17755--17764, 2022.

\bibitem[Vaswani et~al.(2017)Vaswani, Shazeer, Parmar, Uszkoreit, Jones, Gomez, Kaiser, and Polosukhin]{vaswani2017attention}
Ashish Vaswani, Noam Shazeer, Niki Parmar, Jakob Uszkoreit, Llion Jones, Aidan~N Gomez, {\L}ukasz Kaiser, and Illia Polosukhin.
\newblock Attention is all you need.
\newblock \emph{Advances in neural information processing systems}, 30, 2017.

\bibitem[Wang et~al.(2021)Wang, Wang, Dong, Xu, Yang, An, and Guo]{dasr2021wang}
Longguang Wang, Yingqian Wang, Xiaoyu Dong, Qingyu Xu, Jungang Yang, Wei An, and Yulan Guo.
\newblock Unsupervised degradation representation learning for blind super-resolution.
\newblock In \emph{Proceedings of the IEEE/CVF Conference on Computer Vision and Pattern Recognition}, pages 10581--10590, 2021.

\bibitem[Wang et~al.(2004)Wang, Bovik, Sheikh, and Simoncelli]{ssim}
Zhou Wang, Alan~C Bovik, Hamid~R Sheikh, and Eero~P Simoncelli.
\newblock Image quality assessment: from error visibility to structural similarity.
\newblock \emph{IEEE transactions on image processing}, 13\penalty0 (4):\penalty0 600--612, 2004.

\bibitem[Wang et~al.(2022)Wang, Cun, Bao, Zhou, Liu, and Li]{wang2022uformer}
Zhendong Wang, Xiaodong Cun, Jianmin Bao, Wengang Zhou, Jianzhuang Liu, and Houqiang Li.
\newblock Uformer: A general u-shaped transformer for image restoration.
\newblock In \emph{Proceedings of the IEEE/CVF conference on computer vision and pattern recognition}, pages 17683--17693, 2022.

\bibitem[Woo et~al.(2018)Woo, Park, Lee, and Kweon]{woo2018cbam}
Sanghyun Woo, Jongchan Park, Joon-Young Lee, and In~So Kweon.
\newblock Cbam: Convolutional block attention module.
\newblock In \emph{Proceedings of the European conference on computer vision (ECCV)}, pages 3--19, 2018.

\bibitem[Xiao et~al.(2023)Xiao, Fu, Zhou, Liu, and Zha]{shuffleformer}
Jie Xiao, Xueyang Fu, Man Zhou, Hongjian Liu, and Zheng-Jun Zha.
\newblock Random shuffle transformer for image restoration.
\newblock In \emph{International Conference on Machine Learning}, pages 38039--38058. PMLR, 2023.

\bibitem[Xie et~al.(2021)Xie, Wang, Yu, Anandkumar, Alvarez, and Luo]{xie2021segformer}
Enze Xie, Wenhai Wang, Zhiding Yu, Anima Anandkumar, Jose~M Alvarez, and Ping Luo.
\newblock Segformer: Simple and efficient design for semantic segmentation with transformers.
\newblock \emph{Advances in neural information processing systems}, 34:\penalty0 12077--12090, 2021.

\bibitem[Yaqi et~al.(2024)Yaqi, Zhihao, Xiaofeng, Jimmy~S., Xiaoming, Zongsheng, Chongyi, Shangcheng, Ruicheng, Yuekun, Peiqing, Chen~Change, et~al.]{hybridevs2024mipi3}
Wu Yaqi, Fan Zhihao, Chu Xiaofeng, Ren Jimmy~S., Li Xiaoming, Yue Zongsheng, Li Chongyi, Zhou Shangcheng, Feng Ruicheng, Dai Yuekun, Yang Peiqing, Loy Chen~Change, et~al.
\newblock Mipi 2024 challenge on demosaic for hybridevs camera: Methods and results.
\newblock In \emph{Proceedings of the IEEE/CVF Conference on Computer Vision and Pattern Recognition}, 2024.

\bibitem[Yue et~al.(2020)Yue, Zhao, Zhang, and Meng]{yue2020dual}
Zongsheng Yue, Qian Zhao, Lei Zhang, and Deyu Meng.
\newblock Dual adversarial network: Toward real-world noise removal and noise generation.
\newblock In \emph{Computer Vision--ECCV 2020: 16th European Conference, Glasgow, UK, August 23--28, 2020, Proceedings, Part X 16}, pages 41--58. Springer, 2020.

\bibitem[Zamir et~al.(2020{\natexlab{a}})Zamir, Arora, Khan, Hayat, Khan, Yang, and Shao]{mirnet}
Syed~Waqas Zamir, Aditya Arora, Salman Khan, Munawar Hayat, Fahad~Shahbaz Khan, Ming-Hsuan Yang, and Ling Shao.
\newblock Learning enriched features for real image restoration and enhancement.
\newblock In \emph{Computer Vision--ECCV 2020: 16th European Conference, Glasgow, UK, August 23--28, 2020, Proceedings, Part XXV 16}, pages 492--511. Springer, 2020{\natexlab{a}}.

\bibitem[Zamir et~al.(2020{\natexlab{b}})Zamir, Arora, Khan, Hayat, Khan, Yang, and Shao]{zamir2020cycleisp}
Syed~Waqas Zamir, Aditya Arora, Salman Khan, Munawar Hayat, Fahad~Shahbaz Khan, Ming-Hsuan Yang, and Ling Shao.
\newblock Cycleisp: Real image restoration via improved data synthesis.
\newblock In \emph{Proceedings of the IEEE/CVF conference on computer vision and pattern recognition}, pages 2696--2705, 2020{\natexlab{b}}.

\bibitem[Zamir et~al.(2022)Zamir, Arora, Khan, Hayat, Khan, and Yang]{zamir2022restormer}
Syed~Waqas Zamir, Aditya Arora, Salman Khan, Munawar Hayat, Fahad~Shahbaz Khan, and Ming-Hsuan Yang.
\newblock Restormer: Efficient transformer for high-resolution image restoration.
\newblock In \emph{Proceedings of the IEEE/CVF conference on computer vision and pattern recognition}, pages 5728--5739, 2022.

\bibitem[Zhang et~al.(2018{\natexlab{a}})Zhang, Sindagi, and Patel]{zhang2018multi}
He Zhang, Vishwanath Sindagi, and Vishal~M Patel.
\newblock Multi-scale single image dehazing using perceptual pyramid deep network.
\newblock In \emph{Proceedings of the IEEE conference on computer vision and pattern recognition workshops}, pages 902--911, 2018{\natexlab{a}}.

\bibitem[Zhang et~al.(2022)Zhang, Fu, and Li]{demosacicnet2022zhang}
Tao Zhang, Ying Fu, and Cheng Li.
\newblock Deep spatial adaptive network for real image demosaicing.
\newblock In \emph{Proceedings of the AAAI Conference on Artificial Intelligence}, pages 3326--3334, 2022.

\bibitem[Zhang et~al.(2018{\natexlab{b}})Zhang, Li, Li, Wang, Zhong, and Fu]{zhang2018imagesr}
Yulun Zhang, Kunpeng Li, Kai Li, Lichen Wang, Bineng Zhong, and Yun Fu.
\newblock Image super-resolution using very deep residual channel attention networks.
\newblock In \emph{Proceedings of the European conference on computer vision (ECCV)}, pages 286--301, 2018{\natexlab{b}}.

\bibitem[Zheng et~al.(2021)Zheng, Lu, Zhao, Zhu, Luo, Wang, Fu, Feng, Xiang, Torr, et~al.]{zheng2021setr}
Sixiao Zheng, Jiachen Lu, Hengshuang Zhao, Xiatian Zhu, Zekun Luo, Yabiao Wang, Yanwei Fu, Jianfeng Feng, Tao Xiang, Philip~HS Torr, et~al.
\newblock Rethinking semantic segmentation from a sequence-to-sequence perspective with transformers.
\newblock In \emph{Proceedings of the IEEE/CVF conference on computer vision and pattern recognition}, pages 6881--6890, 2021.

\bibitem[Zhou et~al.(2023)Zhou, Li, Guo, Bai, Cheng, and Hou]{zhou2023srformer}
Yupeng Zhou, Zhen Li, Chun-Le Guo, Song Bai, Ming-Ming Cheng, and Qibin Hou.
\newblock Srformer: Permuted self-attention for single image super-resolution.
\newblock In \emph{Proceedings of the IEEE/CVF International Conference on Computer Vision}, pages 12780--12791, 2023.

\bibitem[Zhu et~al.(2020)Zhu, Su, Lu, Li, Wang, and Dai]{zhu2020deformable}
Xizhou Zhu, Weijie Su, Lewei Lu, Bin Li, Xiaogang Wang, and Jifeng Dai.
\newblock Deformable detr: Deformable transformers for end-to-end object detection.
\newblock \emph{arXiv preprint arXiv:2010.04159}, 2020.

\bibitem[Zhu et~al.(2023)Zhu, Wang, Fu, and Hu]{zhu2023enhanced}
Yurui Zhu, Xi Wang, Xueyang Fu, and Xiaowei Hu.
\newblock Enhanced coarse-to-fine network for image restoration from under-display cameras.
\newblock In \emph{Computer Vision--ECCV 2022 Workshops: Tel Aviv, Israel, October 23--27, 2022, Proceedings, Part V}, pages 130--146. Springer, 2023.

\end{thebibliography}
}


\end{document}